\documentclass{article}
\usepackage[square,numbers]{natbib}

\usepackage[preprint]{neurips_2025}

\usepackage[utf8]{inputenc} 
\usepackage[T1]{fontenc}    
\usepackage{hyperref}       
\usepackage{url}            
\usepackage{booktabs}       
\usepackage{amsfonts}       
\usepackage{amsmath}       
\usepackage{nicefrac}       
\usepackage{microtype}      
\usepackage{xcolor}         

\usepackage[inline]{enumitem}
\usepackage{makecell}
\usepackage{fancyhdr}       
\usepackage{graphicx}       
\usepackage{subcaption}
\usepackage{mdframed}
\usepackage{comment}
\usepackage{tabularx}
\usepackage{caption}
\usepackage{placeins}
\usepackage{float}
\usepackage{array} 
\usepackage{wrapfig}
\usepackage{cleveref}
\usepackage{multirow}
\usepackage{algorithm}
\usepackage{algorithmic}
\usepackage{listings}
\usepackage{atbeginend}

\AfterBegin{packed_enum}{\vspace{-0.7em}}
\AfterEnd{packed_enum}{\vspace{-0.7em}}
\AfterBegin{packed_item}{\vspace{-0.7em}}
\AfterEnd{packed_item}{\vspace{-0.7em}}
\AfterBegin{packed_description}{\vspace{-0.7em}}
\AfterEnd{packed_description}{\vspace{-0.7em}}

\AfterBegin{quote}{\vspace{-0.7em}}
\AfterEnd{quote}{\vspace{-0.7em}}

\lstdefinelanguage{json}{
basicstyle=\ttfamily,
numbers=left,
numberstyle=\tiny\color{gray},
stepnumber=1,
numbersep=8pt,
showstringspaces=false,
breaklines=true,
literate=
*{0}{{{\color{blue}0}}}{1}
{1}{{{\color{blue}1}}}{1}
{2}{{{\color{blue}2}}}{1}
{3}{{{\color{blue}3}}}{1}
{4}{{{\color{blue}4}}}{1}
{5}{{{\color{blue}5}}}{1}
{6}{{{\color{blue}6}}}{1}
{7}{{{\color{blue}7}}}{1}
{8}{{{\color{blue}8}}}{1}
{9}{{{\color{blue}9}}}{1}
{:}{{{\color{red}:}}}{1}
{,}{{{\color{red},}}}{1}
{\{}{{{\color{orange}\{}}}{1}
{\}}{{{\color{orange}\}}}}{1}
{[}{{{\color{orange}[}}}{1}
{]}{{{\color{orange}]}}}{1},
}

\lstdefinelanguage{txt}{
basicstyle=\ttfamily,
numbers=left,
numberstyle=\tiny\color{gray},
stepnumber=1,
numbersep=8pt,
showstringspaces=false,
breaklines=true,
}

\usepackage{newfloat}
\floatstyle{ruled}
\newfloat{listing}{tb}{lst}{}
\floatname{listing}{Listing}
\usepackage{authblk}
\usepackage{pgf-umlsd} 
\usepackage{placeins}

\title{Readme\_AI: Dynamic Context Construction for Large Language Models}

\author[1]{Millie Vyas}
\author[2]{Timothy Blattner}
\author[2]{Alden Dima}

\affil[1]{Purdue University}
\affil[2]{National Institute of Standards and Technology}
\affil[ ]{\textit {\{timothy.blattner,alden.dima\}@nist.gov}}

\begin{document}

\maketitle

\begin{abstract}
Despite being trained on significant amounts of data, Large Language Models (LLMs) can provide inaccurate or unreliable information in the context of a user's specific query. Given query-specific context significantly improves the usefulness of its responses. In this paper, we present a specification that can be used to dynamically build context for data sources. The data source owner creates the file containing metadata for LLMs to use when reasoning about dataset-related queries. To demonstrate our proposed specification, we created a prototype Readme\_AI Model Context Protocol (MCP) server that retrieves the metadata from the data source and uses it to dynamically build context. Some features that make this specification dynamic are the extensible types that represent crawling web-pages, fetching data from data repositories, downloading and parsing publications, and general text. The context is formatted and grouped using user-specified tags that provide clear contextual information for the LLM to reason about the content. We demonstrate the capabilities of this early prototype by asking the LLM about the NIST-developed Hedgehog library, for which common LLMs often provides inaccurate and irrelevant responses containing hallucinations. With Readme\_{AI}, the LLM receives enough context that it is now able to reason about the library and its use, and even generate code interpolated from examples that were included in the Readme\_AI file provided by Hedgehog’s developer. Our primary contribution is a extensible protocol for dynamically grounding LLMs in specialized, owner-provided data, enhancing responses from LLMs and reducing hallucinations. The source code for the Readme\_AI tool is posted here: \url{https://github.com/usnistgov/readme_ai}.
\end{abstract}

\section{Introduction}
While Large Language Models (LLMs) generally succeed at generating meaningful responses, they often lack the correct context needed to answer a user’s specific query. This is because the LLM uses the pre-existing training data that is embedded in its weights to predict the average response, which often provides a more probable but inappropriate context to meet user needs. Providing the LLM with the specific appropriate context improves the quality and accuracy of the response and decreases the chances of hallucinations. To address this issue, we developed Readme\_AI, a metadata specification that allows LLMs to build dynamic, query-specific context from external sources. We prototyped a tool that implements this specification using the Model Context Protocol (MCP) and FastMCP library. Data source owners will populate a Readme\_AI.json file with metadata that is useful for LLMs to build enough context to answer a query. If the Readme\_AI.json file exists, then the Readme\_AI tool automatically downloads this metadata from the data source, such as git, parses the JSON file, and processes the file using the Readme\_AI specification to build context. The current specification provides the capability for creating custom tags with descriptions, fetching data from repositories, website crawling, and downloading and parsing PDFs. To demonstrate the capabilities of Readme\_AI, we ask the LLM about the NIST-developed Hedgehog library, a topic that is likely to be underrepresented by the model’s weights, and results in hallucinations when querying about it. The Readme\_AI's JSON file is populated with carefully curated and validated information from the library's owner, including URLs to documentation, example code snippets, API code references from its repository, and parsed publications. This rich context enables the LLM to reason about the library and its usage, ultimately allowing it to generate code that parallelizes execution for specified algorithms. 

The key contributions of our work are:
\begin{enumerate}
    \item An extensible Readme\_AI specification to dynamically build context from up-to-date metadata that is built by the data owner.
    \item A prototype implementation of this protocol.
    \item A demonstration of its ability to reduce hallucination and enable accurate representation from specialized, niche, or rapidly evolving domains.
\end{enumerate}

\section{Related Work}
There are several alternate methods for approaching building context for LLMs to reason about data sources. Retrieval Augmented Generation (RAG) is a common technique to extract context using a vector database built from embedding models~\cite{rag}. This approach relies on the quality of the embedding and identifying similarities between a user’s query and the vector database. RAG is designed to subset a large corpus of data and identify only the most relevant sections. Building an effective RAG relies on significant amounts of tuning, such as data extraction methods and chunking strategies to name a few. Another technique for presenting context to the LLM is Context7~\cite{context7}. Context7 is an MCP server that pulls the latest documentation and code from a given library, extracted from its centralized repository of libraries. The repository is constructed by automatically pulling a variety of documentation files, focusing on building context automatically from git repositories and its existing code snippets. Their specification includes suggestions on how to structure a repository, which could complement the Readme\_AI design. This centralized repository requires developers to remotely connect to Context7's repository to pull in context. Conversely, Readme\_AI's implementation uses a decentralized approach and provides mechanisms for connecting to data sources to download the data source's Readme\_AI JSON file, constructed by the data source owner. Another approach is llms.txt~\cite{llms-txt}, a static file is created by a data source owner that is meant to include relevant information from its data source, presented to the LLM as a markdown file. Elements from this representation is included in the Readme\_AI's specification for its static content. 

Outside of the LLM domain, Frictionless Data provides a Data Package specification~\cite{frictionless_data_package_2017} for describing dataset structure and content to promote interoperability and reusability. Like llms.txt, it creates a static file that provides relevant information about data, but focuses on metadata for human and software consumption such as file schemas, data types, and licenses. 



The Readme\_AI specification is designed to be extensible, allowing for the integration of new context-building methodologies. This aligns with the trend towards modular RAG architectures, but distinguishes itself by empowering data source owners to define and validate the context associated with their data. In the following sections, we introduce the Readme\_AI specification and its abstract grammar. Next, we present an implementation of this specification using the Readme\_AI MCP server. Then, we will demonstrate the approach by building context for a relatively unknown parallel computing library, Hedgehog. Lastly, we will conclude with a discussion on the approach, and ways that we envision that the specification can be extended for vastly more dynamic context building.


\section{Readme\_AI Specification}

The Readme\_AI specification provides a structured format for defining and organizing a data source's metadata within a single JSON file. Its primary goal is to create a machine-readable map of components, which will be transformed into a format that is easily processed by an LLM, such as XML. The specification is implemented in a file named Readme\_AI.json, which should be placed in the root directory of a data source and populated by the owner. 

\subsection{Specification Structure}

The Readme\_AI.json file is a JSON dictionary composed of key-value pairs. Listing~\ref{lst:json-example} presents an example Readme\_AI JSON file that follows the specification. Each key is a string that corresponds to a specific metadata category (e.g., description, files that exist in the data source, documentation). When processed, these keys are used as tags that provide structure for the context sent to the LLM. The value associated with each key can take one of the following forms:
\begin{enumerate}
    \item Simple String: A plain text value, ideal for providing descriptive context for a key.
    \item Structured Object: For complex data, the value can be a general object. Currently containing the following structure:    
        \begin{itemize}
        \item data: A list or dictionary of  strings, typically representing file paths or URLs, followed by a description of the file or URL
        \item type: A string that specifies how the data should be handled (e.g., fetch or crawl)
        \end{itemize}   

        The type key instructs a processing tool on how to handle the structured object. Currently there are three types: (1) fetch, to retrieve files from within a data source like a local repository; (2) crawl, to navigate from a URL and gather all linked files; and (3) download, to directly download files like datasets or PDFs from a list of URLs.
\end{enumerate}

\begin{lstlisting}[language=json, caption={Example Readme\_AI JSON data with Simple string and structured objects}, label={lst:json-example}]
{
  "description": "An example project demonstrating the Readme_AI specification.",
  "source_files": {
    "data": { 
      "/src/main.py": "Main file", 
      "/src/utils.py": "Utility file,
    }
    "type": "fetch"
  },
  "api_files" : {
    "data": [ "/src/api/*" ],
    "type": "fetch"
  },
  "documentation" : {
    "data": "doc-url",
    "type": "crawl"
  }
}
        \end{lstlisting}

Additionally, we have formulated this specification as an abstract grammar, shown in Listing~\ref{lst:grammar}.

\begin{lstlisting}[language=txt, caption={Readme\_AI Abstract Grammar}, label={lst:grammar}]
<readmeai_json> ::= <dict_kv_pair>
<kv_pair> ::= <string>:<data_value>
<data_value> ::= <string>  
               | <structured_object>
<structured_object> ::= {
  "data": <list_string> 
        | <dict_string>, 
  "type": <type_string> 
  }
<type_string> ::= "fetch" | "crawl" | "download"
\end{lstlisting}

\subsection{Extensibility}

The Readme\_AI specification is designed to be flexible and can be extended to meet various project needs. Firstly, top-level keys are arbitrary and serve as tags that provide a structure and organization for the LLM to reason about the tag's content. Additionally, the ``type'' field within a structured object is not limited to predefined values, enabling the creation of custom types and the implementation of custom data handlers within the specification's implementation. Furthermore, the structured object can be extended beyond its current representation, which includes only ``data'' and ``type'' fields, to incorporate additional fields that can express new behaviors or requirements when acquiring context. For example, specifying connections to services or even other tool servers. This would require modification to the underlying server implementation to incorporate these new features. These ideas are discussed further in Section~\ref{sec:future}.



\section{Readme\_AI Implementation}
\begin{figure*}[t]
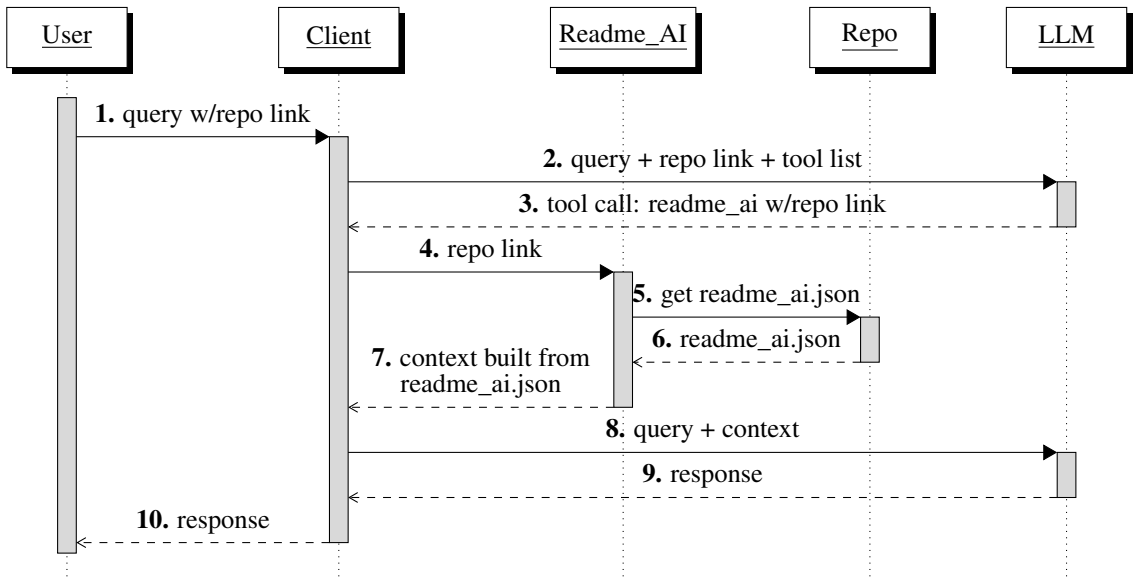

\centering
\makebox[\textwidth][c]{
    \begin{sequencediagram}
        \newthread{u}{User}
        \newinst[2]{c}{Client}
        \newinst[2]{t}{Readme\_AI}
        \newinst[1.5]{r}{Repo}
        \newinst[1]{llm}{LLM}
    
        \begin{call}{u}{\textbf{1.} query w/repo link}{c}{\textbf{10.} response}
            \begin{call}{c}{\textbf{2.} query + repo link + tool list}{llm}{\textbf{3.} tool call: readme\_ai w/repo link}
            \end{call}
            \begin{call}{c}{\textbf{4.} repo link}{t}{\shortstack{\textbf{7.} context built from \\ readme\_ai.json}}
                \begin{call}{t}{\textbf{5.} get readme\_ai.json}{r}{\textbf{6.} readme\_ai.json}
                \end{call}
            \end{call}
            \begin{call}{c}{\textbf{8.} query + context}{llm}{\textbf{9.} response}
            \end{call}
        \end{call}
    \end{sequencediagram}
}
    \caption{Sequence diagram illustrating the Readme\_AI tool's workflow}
    \label{fig:readme_ai_workflow}
\end{figure*}

To demonstrate the Readme\_AI specification, we implemented a Readme\_AI tool through a Model Context Protocol (MCP) \cite{modelcontextprotocol} server using Python and the FastMCP library \cite{fastmcp}. MCP is a recent advancement in generalized tool usage for LLMs. The details of how a client and LLM interfaces with the Readme\_AI MCP server is described in Figure~\ref{fig:readme_ai_workflow}. 


First, a user submits a query to a client. Because the client has the Readme\_AI tool available, it constructs a prompt that includes the original user's query along with schemas of each tool, which is passed to the LLM. 
A snippet of the schema for the Readme\_AI tool is found in Listing~\ref{lst:tool_description}, with a complete description found in the Readme\_AI repository: \url{https://github.com/usnistgov/readme_ai}. The LLM then decides whether to call the tool or not along with any parameters generalized from the tool's schema. The client is responsible for executing the tool call. As shown in the Readme\_AI's tool description, in order to select the correct data source, the user must provide either a name of a data source that already has a URL registered or give the URL to the data source directly in the user's query. By using either the name or URL for a data source, our tool has enough information to: (1) download the latest version of the Readme\_AI.json file from the base URL, (2) parse the JSON file following the Readme\_AI specification, and (3) process all top-level keys to build context. The context is then returned to the client, which, alongside the original query, is sent back to the LLM to generate a response. With the new context, the LLM is more grounded in the data source's context.

\FloatBarrier

 \begin{lstlisting}[language=txt, caption={Snippet of Readme\_AI Tool Description}, label={lst:tool_description}]
The function is designed to generate comprehensive context about a specified data source repository. This context is crucial for enhancing the LLM's understanding and ability to provide accurate and relevant information related to the data source.
Input Parameters: 
    - URL or Library Name: ...
Functionality: 
Upon receiving the input (URL or library name), the repository will be cloned or pulled for updates, and the Readme_AI.json file will be accessed. The function leverages pre-created context by the data source owner to build a rich understanding of the data source. 
The context includes:
    - Files of Interest: ...
Examples:
To illustrate the function's utility, consider the following examples:
Input: The URL or library name of
a GitHub repository for the NIST 
developed Hedgehog library.

    
\end{lstlisting}

We implemented three functions to handle the fetch, crawl, and download operations for structured objects. The correct function is selected based on the object's ``type'' key.

\begin{description}
    \item[Download] \hfill \\ The function get\_relevant\_papers is called, which takes the PDF URLs, downloads each them, and extracts its contents using the PyPDF Python library~\cite{pypdf}.
    \item[Fetch] \hfill \\ The function fetch\_data is called to access the data source's repository, and include files specified. This method uses relative paths to the files based on the repository's base name, and is implemented using the git Python library~\cite{GitPython}.
    \item[Crawl] \hfill \\ The function web\_crawler is called crawl a list of URLs using the CRAWL package~\cite{BeautifulSoup4} and iteratively builds context as it follows the links across the web pages available from the base URL. 
\end{description}

When a user provides a URL, the tool saves it along with the library's name in a ``lookup.json'' file. This allows the user to simply use the name for subsequent queries instead of re-entering the entire link.

While the Readme\_AI tool currently only supports Git repositories, we plan to extend its functionality to other data sources in the future.

\section{Readme\_AI Example}
To demonstrate Readme\_AI's capabilities, we configured a Readme\_AI.json file for the NIST Hedgehog library~\cite{hedgehog2022} and designed two test queries: one to showcase a basic understanding of Hedgehog and another to illustrate the library's implementation. The Hedgehog library itself uses a structured dataflow-based approach to enable parallel execution in C++. 

\subsection{Basic Chat Example}
To test the LLM's basic understanding, we have developed a test query shown in Listing~\ref{lst:hedgehog_query}.

\begin{lstlisting}[language=txt,caption={Hedgehog Query},label={lst:hedgehog_query}]
Can you describe in a short one sentence summary to me about what the Hedgehog library is that was developed at the National Institute of Standards and Technology, found in: https://github.com/usnistgov/hedgehog
\end{lstlisting}

The response generated by the Llama~4 Maverick FP8 model~\cite{meta_llama4_2025}, shown in Listing~\ref{lst:hedgehog_answer_no_tool}, indicates that the LLM is unaware of the library and its purpose.

\begin{lstlisting}[language=txt,caption={Hedgehog Query Result Without Readme\_AI Tool, Hallucination},label={lst:hedgehog_answer_no_tool}]
The Hedgehog library is a C++ software framework developed by the National Institute of Standards and Technology (NIST) that provides a modular and extensible platform for implementing and testing various numerical methods for solving partial differential equations (PDEs).
\end{lstlisting}

Next, we queried the LLM again with the Readme\_AI tool enabled. When the same query is passed in, the client follows the workflow from Figure~\ref{fig:readme_ai_workflow}, resulting in the Readme\_AI tool being called with the git repository URL as a parameter. The tool then fetched the repository, processed its Readme\_AI.json file, and generated XML-formatted context for the model, shown in Listing~\ref{lst:hedgehog_example}. The XML format was chosen as it produces structured representations of the tags and data extracted from the JSON. Alternate representations such as Markdown could be used as well. Given the new context the LLM responds with a more grounded answer, as shown in Listing~\ref{lst:hedgehog_answer_with_tool}. The LLM extracted the description field from the output of the Readme\_AI tool to answer the question.



\begin{lstlisting}[language=txt, caption={Partial Readme\_AI Hedgehog Library Output}, label={lst:hedgehog_example}]
<DESCRIPTION>
    The Hedgehog library is a C++ 
    library developed by NIST for 
    creating parallel computations
    on heterogeneous nodes.
</DESCRIPTION>
<API>
    <DESCRIPTION>
        This is all the API code for 
        the Hedgehog library ...
    <\DESCRIPTION>

    <file1>        
        AbstractExecutionPipeline ...  
    </file1>    
<\API>
<PAPERS>
    <paper1>        
        This paper presents Hedgehog ...
    <\paper1>    
<\PAPERS>
<WEBSITES>
    <link1>        
        The goal of this tutorial is ...
    <\link1>
<\WEBSITES>
\end{lstlisting}

\begin{lstlisting}[language=txt,caption={Hedgehog Query Result With Readme\_AI Tool},label={lst:hedgehog_answer_with_tool}]
The Hedgehog library is a C++ library developed by NIST for creating parallel computations on heterogeneous nodes.
\end{lstlisting}

One current limitation of this approach is that it can easily generate large contexts. For example, Hedgehog's Readme\_AI JSON output contains approximately 97,240 tokens, which can lead to increased costs when using cloud services. To mitigate this, we plan to integrate an agentic approach that leverages the LLM to identify relevant sections within the JSON output. This could involve querying the LLM to determine the most relevant tags and only including those in the output. While this optimization would reduce token costs, it risks omitting important information and requires data owners to design meaningful tags that the LLM can effectively reason about.

\subsection{Coding Example}

To test if the LLM fully grasped all the context we provided, we prompted the LLM to generate a particle simulator algorithm running in parallel using the Hedgehog library. Without the Readme\_AI tool, the code generated by the LLM hallucinated a Hedgehog library that does not exist, shown in Listing~\ref{lst:particle_sim_without_tool}.  After providing the LLM with context about the NIST-developed Hedgehog library using Readme\_AI, the LLM was able to reference the Hedgehog API code and produce a working particle simulator running in parallel, shown in Listing~\ref{lst:particle_sim_with_tool}. This indicates that the LLM is aware of the Hedgehog library and how to use it. This is accomplished by crawling the documentation and presenting the API codebase to the LLM. Even though there is no current example of how to implement a particle simulator, the documentation provides enough context for the LLM to generalize implementing using the library.

\begin{lstlisting}[language=json, caption={Snippet of Particle Simulator Code Without Tool}, label={lst:particle_sim_without_tool}]
import numpy as np
from hedgehog import Hedgehog
...
class ParticleSimulator:
    def __init__(self, num_particles):
        ...
        self.hedgehog = Hedgehog()
    ...
    # Use Hedgehog for parallel processing
        updated_positions, updated_velocities = self.hedgehog.parallel_update(positions, velocities)
    ...
\end{lstlisting}

\begin{lstlisting}[language=json, caption={Snippet of Particle Simulator Code With Tool}, label={lst:particle_sim_with_tool}]
#include <SFML/Graphics.hpp>
#include <hedgehog/hedgehog.h>
#include <random>
#include <vector>

...

class ParticleSimulationTask : public hh::AbstractTask<1, Block, Block> {
...
    void execute(std::shared_ptr<Block> block, int width, int height) {
        ...
    }
    std::shared_ptr<hh::AbstractTask<1, Block, Block>> copy() {
        ...
}
...
    auto graph = std::make_shared<hh::Graph<1, std::vector<Particle>, Block>>();
...
\end{lstlisting}

\section{Discussion}
The Readme AI tool has demonstrated a significant improvement in the quality and accuracy of the LLM's responses by providing dynamic and query-specific context. A key strength of this approach is that the data source owner is responsible for populating the Readme\_AI.json file, ensuring that the information included is validated and accurate. Without the tool, the model's answer about the NIST-developed Hedgehog library was largely hallucinated and incorrect. While it did talk about a ``Hedgehog'' library, it thought that the library is used for solving partial differential equations. After populating the Readme\_AI.json file with information about the NIST Hedgehog library, asking the LLM about Hedgehog using the tool provided a well-detailed and reliable response. It was able to retain all the information and inform the user that Hedgehog is a C++ library developed by NIST and is used for parallel computations. To see if the LLM built a deeper understanding of Hedgehog, we prompted it to generate code for a parallel Particle Simulator. The task took several attempts, but after we gave the LLM a clear understanding of the goal, it achieves something it was not able to do without the tool. The tool's capability to fetch data from various sources, including repositories, websites, and publications, and to structure this information using user-defined tags, has been particularly effective in enhancing the LLM's understanding. Through this testing, we can conclude that dynamically building context using the Readme\_AI tool significantly improves the quality of the LLM’s response. The capabilities of this tool relies on the construction of well curated data that is incorporated into the Readme\_AI json file.

Readme\_AI introduces a novel methodology for dynamic context building. Unlike traditional RAG implementations that often rely on centralized and pre-processed knowledge bases, Readme\_AI's innovation lies in its specification for dynamically fetching and integrating data from user-curated sources. This enables fetching real-time data such as documentation, code, text, websites, and papers, directly from their latest sources, thus preventing the context collapse that can occur with outdated, static information.

\section{Future Work}
\label{sec:future}
Further refining the Readme\_AI tool involves allowing the LLM to build context from datasets. Directly incorporating Frictionless Data Package specifications into the Readme\_AI context-building process would allow the LLM to not only understand the narrative and functional aspects of a data source through documentation and code but also to gain a structured, machine-readable understanding of datasets themselves. This could be implemented by including a new ``type'' for handling Data Packages. 

The inclusion of Data Packages will allow the data source owner to provide structured data directly as context for the LLM. The inclusion of this capability will have a remarkable impact on LLM performance. Some benefits include performing more sophisticated data analysis, identifying trends in data, and generating code for data visualization. Providing direct access to relevant datasets would enable LLMs to provide highly accurate insights that are beyond their general training knowledge. We also plan to update the JSON specification to incorporate more advanced types such as Retrieval-Augmented Generation (RAG). This will further enable the Readme\_AI tool to capture even more dynamic context building. 

Another feature that is needed is subsetting the context from Readme\_AI to not explode the context with too much content. One strategy that we are contemplating is encouraging well-defined top-level keys that can be used to subset what portions of the json should be emitted for the LLM, thereby reducing the number of tokens used for context and improving the efficiency of the LLM's processing. Additionally, we aim to extend the tool's capability to recursively link to other data sources, pulling their Readme\_AI.json files and building context from these sources as well.

Lastly, an analysis on security controls related to this approach will be needed. As dynamic context is being fetched from open sources, it is possible that backdoor attacks can become more prevalent. One possible technique that can be employed is to specify approved/blocked lists for data sources within the Readme\_AI MCP server. This will ensure that only validated domains will gather data. In general, these security concerns will only grow as more and more services leverage MCP servers.

\section*{Acknowledgments}
The authors utilized Meta's Llama-4-Maverick-17B-128E-Instruct and Google's Gemini 2.5 Pro language models to generate Python and C++ source code and edit the manuscript for clarity and conciseness.

\section*{Disclaimer}
\begin{footnotesize}
No approval or endorsement of any commercial product by the National Institute of Standards and Technology is intended or implied. Certain commercial software, products, and systems are identified in this report to facilitate better understanding. Such identification does not imply recommendations or endorsement by NIST, nor does it imply that the software and products identified are necessarily the best available for the purpose.
\end{footnotesize}


\newpage
\bibliographystyle{IEEEtran}
{
\small
\bibliography{references}

\begin{thebibliography}{10}
\providecommand{\url}[1]{#1}
\csname url@samestyle\endcsname
\providecommand{\newblock}{\relax}
\providecommand{\bibinfo}[2]{#2}
\providecommand{\BIBentrySTDinterwordspacing}{\spaceskip=0pt\relax}
\providecommand{\BIBentryALTinterwordstretchfactor}{4}
\providecommand{\BIBentryALTinterwordspacing}{\spaceskip=\fontdimen2\font plus
\BIBentryALTinterwordstretchfactor\fontdimen3\font minus
  \fontdimen4\font\relax}
\providecommand{\BIBforeignlanguage}[2]{{%
\expandafter\ifx\csname l@#1\endcsname\relax
\typeout{** WARNING: IEEEtran.bst: No hyphenation pattern has been}%
\typeout{** loaded for the language `#1'. Using the pattern for}%
\typeout{** the default language instead.}%
\else
\language=\csname l@#1\endcsname
\fi
#2}}
\providecommand{\BIBdecl}{\relax}
\BIBdecl

\bibitem{rag}
\BIBentryALTinterwordspacing
P.~Lewis, E.~Perez, A.~Piktus, F.~Petroni, V.~Karpukhin, N.~Goyal, H.~Küttler,
  M.~Lewis, W.~tau Yih, T.~Rocktäschel, S.~Riedel, and D.~Kiela,
  ``Retrieval-augmented generation for knowledge-intensive nlp tasks,'' 2021.
  [Online]. Available: \url{https://arxiv.org/abs/2005.11401}
\BIBentrySTDinterwordspacing

\bibitem{context7}
upstash, ``Context7 mcp server,'' \url{https://github.com/upstash/context7},
  accessed: 2025, gitHub repository.

\bibitem{llms-txt}
thedaviddias, ``llms.txt,'' \url{https://github.com/llms-txt/llms-txt},
  accessed: 2025, gitHub repository.

\bibitem{frictionless_data_package_2017}
\BIBentryALTinterwordspacing
R.~P. Paul~Walsh, ``Data package,'' Specification, 2017. [Online]. Available:
  \url{https://specs.frictionlessdata.io/data-package/}
\BIBentrySTDinterwordspacing

\bibitem{modelcontextprotocol}
\BIBentryALTinterwordspacing
``Model context protocol,'' accessed: 2025, gitHub repository. [Online].
  Available: \url{https://github.com/modelcontextprotocol}
\BIBentrySTDinterwordspacing

\bibitem{fastmcp}
\BIBentryALTinterwordspacing
jlowin, ``Fastmcp,'' accessed: 2025. [Online]. Available:
  \url{https://github.com/jlowin/fastmcp}
\BIBentrySTDinterwordspacing

\bibitem{pypdf}
\BIBentryALTinterwordspacing
py~pdf, ``pypdf,'' 2005, gitHub Repository. [Online]. Available:
  \url{https://github.com/py-pdf/pypdf}
\BIBentrySTDinterwordspacing

\bibitem{GitPython}
\BIBentryALTinterwordspacing
{gitpython-developers}, ``{GitPython},'' 2009, gitHub Repository. [Online].
  Available: \url{https://github.com/gitpython-developers/GitPython}
\BIBentrySTDinterwordspacing

\bibitem{BeautifulSoup4}
\BIBentryALTinterwordspacing
L.~Richardson and {BeautifulSoup community}, ``{BeautifulSoup4},'' accessed:
  2025, gitHub Repository. [Online]. Available:
  \url{https://github.com/BeautifulSoup/BeautifulSoup4}
\BIBentrySTDinterwordspacing

\bibitem{hedgehog2022}
\BIBentryALTinterwordspacing
T.~Blattner, A.~Bardakoff, and W.~Keyrouz, ``hedgehog,'' 2022. [Online].
  Available: \url{https://github.com/usnistgov/hedgehog}
\BIBentrySTDinterwordspacing

\bibitem{meta_llama4_2025}
\BIBentryALTinterwordspacing
{Meta AI}, ``The llama 4 herd: The beginning of a new era of natively
  multimodal ai innovation,'' Blog post, April 2025. [Online]. Available:
  \url{https://ai.meta.com/blog/llama-4-multimodal-intelligence/}
\BIBentrySTDinterwordspacing

\end{thebibliography}
}

\end{document}